# IT-Dendrogram: A New Member of the In-Tree (IT) Clustering Family

**Teng Qiu**

(qiutengcool@163.com)

**Yongjie Li**

(liyj@uestc.edu.cn)
University of Electronic Science and
Technology of China, Chengdu, China
*Corresponding author.

**Abstract**: Previously, we proposed a physically-inspired method to construct data points into an effective in-tree (IT) structure, in which the underlying cluster structure in the dataset is well revealed. Although there are some edges in the IT structure requiring to be removed, such undesired edges are generally distinguishable from other edges and thus are easy to be determined. For instance, when the IT structures for the 2-dimensional (2D) datasets are graphically presented, those undesired edges can be easily spotted and interactively determined. However, in practice, there are many datasets that do not lie in the 2D Euclidean space, thus their IT structures cannot be graphically presented. But if we can effectively map those IT structures into a visualized space in which the salient features of those undesired edges are preserved, then the undesired edges in the IT structures can still be visually determined in a visualization environment. Previously, this purpose was reached by our method called *IT-map*. The outstanding advantage of IT-map is that clusters can still be found even with the so-called crowding problem in the embedding. In this paper, we propose another method, called *IT-Dendrogram*, to achieve the same goal through an effective combination of the IT structure and the single link hierarchical clustering (SLHC) method. Like IT-map, IT-Dendrogram can also effectively represent the IT structures in a visualization environment, whereas using another form, called the *Dendrogram*. IT-Dendrogram can serve as another visualization method to determine the undesired edges in the IT structures and thus benefit the IT-based clustering analysis. This was demonstrated on several datasets with different shapes, dimensions, and attributes. Unlike IT-map, IT-Dendrogram can always avoid the crowding problem, which could help users make more reliable cluster analysis in certain problems.

## 1 Introduction
### 1.1 physically inspired in-tree (IT) structure

Previously (1), we proposed a physically-inspired nearest neighbor descent (NND) method to construct data points into an effective in-tree (IT) structure, which proves to be an effective and useful structure especially in cluster analysis, since this physically-inspired IT structure can clearly reveal the cluster structure underlying the dataset, as Fig. 1 shows. Although this IT structure is imperfect that some undesired edges in it require to be removed (e.g., the edge between the two elongated clusters in Fig. 1), such undesired edges are usually distinguishable from other edges and thus are not hard to be determined. The comparison in (2) between this IT structure and common graphs, such as the *k*-nearest-neighbor graph (*k*-NN) and the minimal spanning tree (MST) (3), etc., further proves the saliency of the undesired edges in the

IT structure. After removing those undesired edges, the IT structure will be divided into several independent sub-graphs. each representing one cluster.

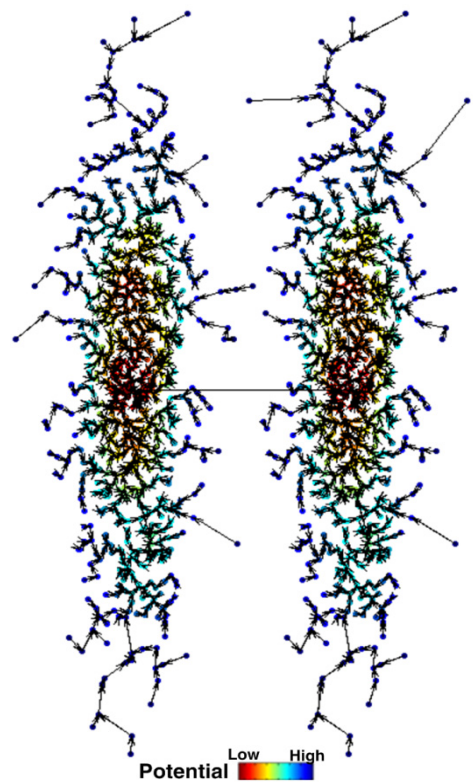

**Fig. 1. The physically-inspired IT structures for a dataset with two elongated clusters.** The colors on points denote the potential values on the points.

**1.2 Visualization (or interactive) methods to remove the undesired edges in IT**

Besides the automatic methods, e.g. G-AP (4) or the Semi-supervised method (denoted as IT-SS here) (1), we are always of the greatest interest to propose the visualization methods to remove those undesired edges in the IT structure, such as **IT-Int** (1), **IT-map** (5) and **IT-DC** (1) , which are detailed as follows.

**IT-Int**. As shown in Fig. 1, when the test dataset is visible in the 2D Euclidean space, the IT structure constructed for the dataset can be visualized and the undesired edge(s) in the IT structure can be spotted effortlessly and thus can be removed interactively. See (1) for details of this interactive cutting method (denoted as *IT-Int* here). This interactive way can also be of a broad meaning, since when the data points are invisible (not in the 2D Eulidean space), they can be first mapped into the 2D Euclidean by one of many linear, e.g., PCA (6), MDS (7), or nonlinear dimensionality reduction methods, e.g., Isomap (8), LLE (9), etc. After that, IT-Int still works. However, the dimensionality methods are not always reliable due to the so-called "crowding problem" (10), that is, data points from different clusters could be overlapped in the embedding, especially when the intrinsic dimensionality in the dataset is higher than 2. This crowding problem makes it hard for clustering methods to distinguish those overlapped clusters in the embedding.

**IT-map**. In (5), we thus proposed a visualization method, call *IT-map*, which no longer requires the dimensionality reduction methods. In fact, to some degree, IT-Map could also be viewed as a dimensionality reduction method. However, unlike the previous dimensionality methods, (i) IT-map directly maps the whole IT structure, i.e., not only the nodes but also the connection relationships between the nodes, into the 2D Euclidean space, in which the undesired edges can be visualized and determined interactively; (ii) IT-map is not affected by the crowding problem in terms of cluster analysis, which should own to the fact that the connections between the nodes in the IT are preserved while mapping, and thus those overlapped nodes can still be assigned into their true clusters respectively.

**IT-DC**. In (1), we also showed the amazing relationships between our physically inspired NND method with Rodriguez and Laio's "Decision Graph" (DG) clustering method (11). Although we were of different origination and angle, we reached almost the same destination. Our method could be roughly* viewed as a graph implementation of their method, if the major difference about the root node is ignored. In turn, their DG can also be derived from our IT structure. Roughly speaking, their DG and out IT structure are to some degree equivalent in application. In fact, DG could be viewed as another efficient visualization method to determine those undesired edges in our IT structures (Here we denote this method as *IT-DC*). After mapping all the nodes in IT structure into a 2D scatter plot featured with the potential and edge length variables in the IT structure, the start nodes of those undesired edges in the IT structure would be distinguishable from the other nodes and thus can be interactively determined. Like IT-map, IT-DC does not need the dimensionality reduction methods, either.

Now, we would like to show to the readers another effective visualization method, called **IT-Dendrogram**, for which dimensionality reduction methods are also unnecessary.

**2 Motivation**

Since it is known that the minimal spanning tree (MST) generally corresponds to the *Dendrogram* obtained by the single link hierarchical clustering method (see Section 3.1 for details), there is hope that we can represent our IT data structure by the Dendrogram, since the IT structure also belongs to the tree structure.

**3 Method**
**3.1 Preliminary information for hierarchical clustering and Dendrogram**
**3.1.1 Hierarchical clustering (HC)**

Hierarchical clustering (HC) methods are widely used in the fields as biology, social science, medicine, archaeology, computer science and engineering (12). There are two categories for (HC) methods: agglomerative and divisive. Agglomerative HC methods first take each data instance as a cluster, then they successively merge two clusters with the smallest dissimilarity as one cluster, until all the data instances are contained in one cluster. The divisive methods take the inverse process. Here, we mainly concern the agglomerative HC methods. The main problem for HC methods is that they are time-consuming. Take the agglomerative methods for example. Many computations are needed to measure the dissimilarities between all clusters so as to determine which are the two clusters with the smallest dissimilarity that ought to be merged each time.

### 3.1.2 Dendrogram

The advantage for HC methods should be due to the fact that the whole clustering process can be visualized via the Dendrogram (or called the hierarchical tree, as shown in Fig. 2B), which makes the HC methods quite popular especially in bioinformatics (13). In the Dendrogram, all data points are represented by the leaf nodes in the bottom layer; each merging process is denoted by a "∩"-shaped link, with its two vertical lines connecting the two merged clusters and the height (y-axis) of its horizontal line denotes the dissimilarity between the merged clusters. As the merging proceeds, this hierarchical tree is nested layer by layer from bottom to up. In the Dendrogram, some of the "∩"-shaped links could be relatively higher, which reveals that the clusters linked by them have larger dissimilarities and thus should not be merged together. By cutting off the links at certain dissimilarity level or threshold, one can obtain several independent sub-dendrograms, each representing one cluster. The leaf nodes in the same sub-dendrograms are thus assigned in the same clusters. One advantage for the Dendrogram is that all data instances (or the leaf nodes) are explicitly arranged without overlapping.

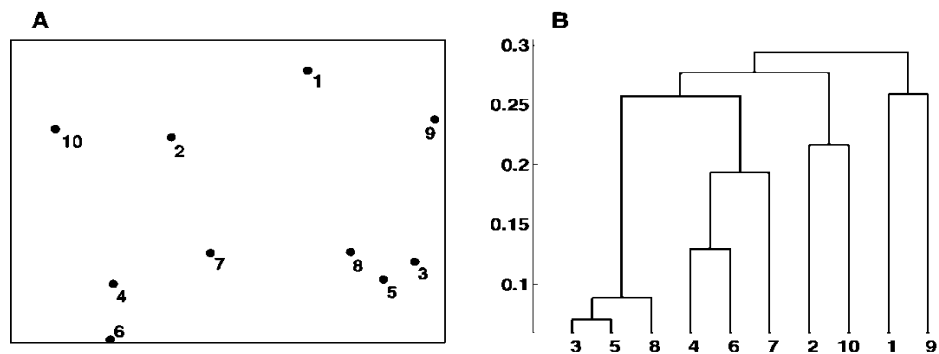

**Fig. 2. An illustration for the SLHC method and Dendrogram.** (A) the test dataset, with 10 data points, the indexes of which are attached aside. (B) The Dendrogram corresponding to the SLHC data structure of the input datasets in (A). The leaf nodes (bottom layer) correspond to the data points in (A), with the corresponding data indexes attached beneath them. The vertical axis shows the dissimilarities of the merged clusters (or the heights of the "∩"-shaped links).

### 3.1.3 Single link hierarchical clustering (SLHC)

For measuring the dissimilarities between clusters, there are a set of methods available, each having its own superiority in certain problems. One representative, called the single link (SL) algorithm, defines the dissimilarity between clusters $C_i$ and $C_j$ as the smallest one among the dissimilarities between all pair of data instances respectively from clusters $C_i$ and $C_j$. We denote here the agglomerative HC based on SL algorithm as SLHC. Besides, in order to measure the dissimilarity between data instances, one also needs to choose the measurement, e.g., the Euclidean distance.

### 3.1.4 Relationships between SLHC and MST algorithm

One interesting thing for SLHC is that, it has a very close relationship with the

MST algorithm. The dissimilarities (from start to end) for all pairs of merged clusters or the heights (from bottom to top) of the "∩"-shaped links in the Dendrogram obtained by SLHC method correspond to the edge dissimilarities (in ascending order) in the MST of the same datasets, which is true at least when the dissimilarities between all data instances are different (12).

**3.2 The proposed method: IT-Dendrogram**

The visualization goal of representing the IT data structure by the Dendrogram in a visualization environment is achieved by an effective combination of the IT data structure with SLHC. In short, we let SLHC method take as input the IT data structure of the test datasets instead of the test datasets directly. See details as follows.

**Step 0:** transform the input dissimilarity matrix D to the IT data structure by using the nearest neighbor descent (NND) algorithm[1] (*1*) or a generalized NND (G-NND) (*2*). The IT data structure is represented by three vectors *I*, *W* and *P*. See detailed computation in (*1*). $I_i$ denotes the node that node *i* "descends" to, and $W_i$ denotes the dissimilarity between them (i.e., nodes *i* and $I_i$). $P_i$ denotes the potential associated with each node *i*. Note that there is a parameter (denoted as *σ*) in this step. We treat this step as a preliminary step[2], since it is our previous work of constructing the IT data structure.

**Step 1**: transform the IT data structure (ignore the potential vector *P*) to a sparse matrix *S*, for which $S_{i,I_i} = S_{I_i,i} = W_i$.

**Step 2:** transform the matrix *S* to the single-link based hierarchical clustering data structure *Z* by SLHC method. In MATLAB, one can use the "linkage" function[3] to obtain the matrix *Z*, with (*N-1*) rows and 3 columns, in which $Z_{i1}$ and $Z_{i2}$ record the indexes[4] of the merged clusters in the *i*-th time (or layer), and $Z_{i3}$ records the dissimilarity of the merged clusters $Z_{i1}$ and $Z_{i2}$.

**Step 3**: plot the Dendrogram corresponding to the matrix *Z*. In MATLAB, one can use the "Dendrogram" function to obtain the Dendrogram.

From the steps above, one can see that compared with IT-map which uses all information (i.e., vectors *I*, *W* and *P*) in the IT data structure, IT-Dendrogram uses incomplete information (i.e., vectors *I* and *W* only) to achieve the same goal of visualizing the IT structures.

**3.3 Fast IT-Dendrogram**

In fact, it is unnecessary to run the time-consuming SLHC method to obtain the single-link based hierarchical data structure *Z*. One can use one simple and direct step to replace steps 1 and 2 above. Given the fact introduced in the Section 3.1.4, the third column of *Z* is actually the edge dissimilarities *W* (in ascending order) in the IT structure. In other words, the dissimilarities of all the merged clusters in the merging process computed by SLHC are known in advance. $Z_{i1}$ and $Z_{i2}$ actually denote the clusters where the start and end nodes of the directed edge in the IT data structure with the edge length $Z_{i3}$ come from. As $Z_{i3}$ is known, $Z_{i1}$ and $Z_{i2}$ are thus easy to be identified with reference to the index vector $I$ accompanied.

---

[1] The experiments in Section 4 were based on the NND algorithm.
[2] That is why we call it step 0. Strictly speaking, this step is not included in IT-Dendrogram.
[3] Z = linkage (S, 'single').
[4] Note that, at the beginning, each data instance is viewed as a cluster. The indexes of them are among 1, 2, ..., N. As the merging proceed, the index of each newly generated cluster is accumulated on the max index (denoted as M) of the previous clusters, i.e., M+1.

### 3.4 The rationality behind IT-Dendrogram

We can describe the content in Section 3.1.4 in a more general way, that is, when SLHC method takes as input the dissimilarity matrix corresponding to graph $G$, then the obtained Dendrogram should correspond to the MST derived from graph $G$. When $G$ is the IT' (here we denote the IT ignoring the edge direction as IT'), then the obtained Dendrogram should correspond to the MST derived from the IT'. It is obvious that the MST structure derived from a tree structure, saying the IT', is the IT' itself. Since, according to the definition for the MST—the spanning tree with the least sum of edge lengths among all spanning trees of the original graph, there is only one spanning tree for the IT', namely itself, therefore the derived MST from the IT' is the IT' itself. In other words, the Dendrogram constructed by the proposed method is an effective mapping for the IT' or the IT.

### 3.5 Cluster analysis based on IT-Dendrogram

From the Dendrogram, we can determine the threshold to separate the inconsistent links (i.e., those "∩"-shaped links with relative high dissimilarities) from the other links. After the threshold in the Dendrogram is determined, the threshold for separating the undesired edges (i.e., the edges with larger dissimilarities or edge lengths than the threshold) from the other edges in the IT structure is also determined, since these two thresholds are the same (see Sections 3.3 and 3.4). After removing the undesired edges in the IT structure based on the threshold, the clustering assignment can be performed by the root-finding step in our IT-based clustering framework (1). The data instances with the same root nodes will be assigned in the same clusters.

Of course, based on the Dendrograms, the clustering assignment can be performed in the framework of the SLHC method by searching for the leaf nodes (see Section 3.1.2), whereas our root-finding process should be more efficient.

### 4 Experiments

We tested the proposed method IT-Dendrogram on several synthetic datasets with different shapes, dimensions and attributes, as listed in Table 1.

Table 1. The test datasets

| Dataset | Attribute | $d$ | $N$ |
|---|---|---|---|
| S4 (2) | real | 2 | 5000 |
| Science-14 (14) | real | 2 | 4000 |
| Aggregation (11) | real | 2 | 788 |
| Flame (15) | real | 2 | 240 |
| Spiral (16) | real | 2 | 600 |
| D32 (17) | real | 32 | 1024 |
| Mushroom | Char | 22 | 8124 |

*Attribute: the attribute of the elements in each data instance (or vector) in the dataset; $d$: dimensionality of the data instance. $N$: number of the data instance.

The first five datasets in Table 1 are 2D synthetic datasets[5] with varying numbers of real-valued data points (or data instances or vectors). Their Dendrograms constructed by IT-Dendrogram are shown in the 1st (leftmost) column of Fig. 3 (From up to bottom, parameter $\sigma$ = 10000, 1, 0.05, 1.8, 2 in step 0). We can see that several clusters are merged by the "∩"-shaped links with relative high dissimilarities than the other cases. The thresholds (denoted as the heights of the red dashed lines) can be used to separate those inconsistent "∩"-shaped links from the rest ones. As stated in Section 3.5, these thresholds are used to cut the undesired edges in the IT structure. Note that for visualization purpose, here we also used these thresholds to cut those Dendrograms and used different colors to represent different sub-Dendrograms, whereas these operations are not necessary in practice[6]. The corresponding clustering results after removing the undesired edges in the IT structures are shown in the 2nd column in Fig. 3. We can see that all these results are very consistent with the underlying cluster structures perceived by us.

We also made a comparison with SLHC method. The Dendrograms and the corresponding clustering results are shown in the 3rd and 4th columns in Fig. 3, respectively. We can see that SLHC only works well on the last dataset ("Spiral dataset" in the bottom), whereas quite poorly on the other datasets.

The reason for the superiority of our IT-Dendrogram over SLHC is due to the fact given in Section 3.4, that is, the Dendrograms obtained by SLHC generally correspond to the MST data structures of the test datasets while the Dendrograms obtained by IT-Dendrogram correspond to the physically-inspired IT structures of the test datasets, and generally, the undesired edges in the IT structures are much more salient or distinguishable than those in the MST structures, as Fig. 4 shows. In other words, this superiority is in nature due to the superiority of the IT structures over the MST structures in cluster analysis, as also discussed in (18).

---

[5] Download from http://cs.joensuu.fi/sipu/datasets/
[6] For example, in Fig. 5, we didn't cut the Dendrogram, since the main function for Dendrogram is to provide us a visualized reference to judge how well the underlying IT structures are constructed and help us to set a better threshold to cut the undesired edges in the IT structures.

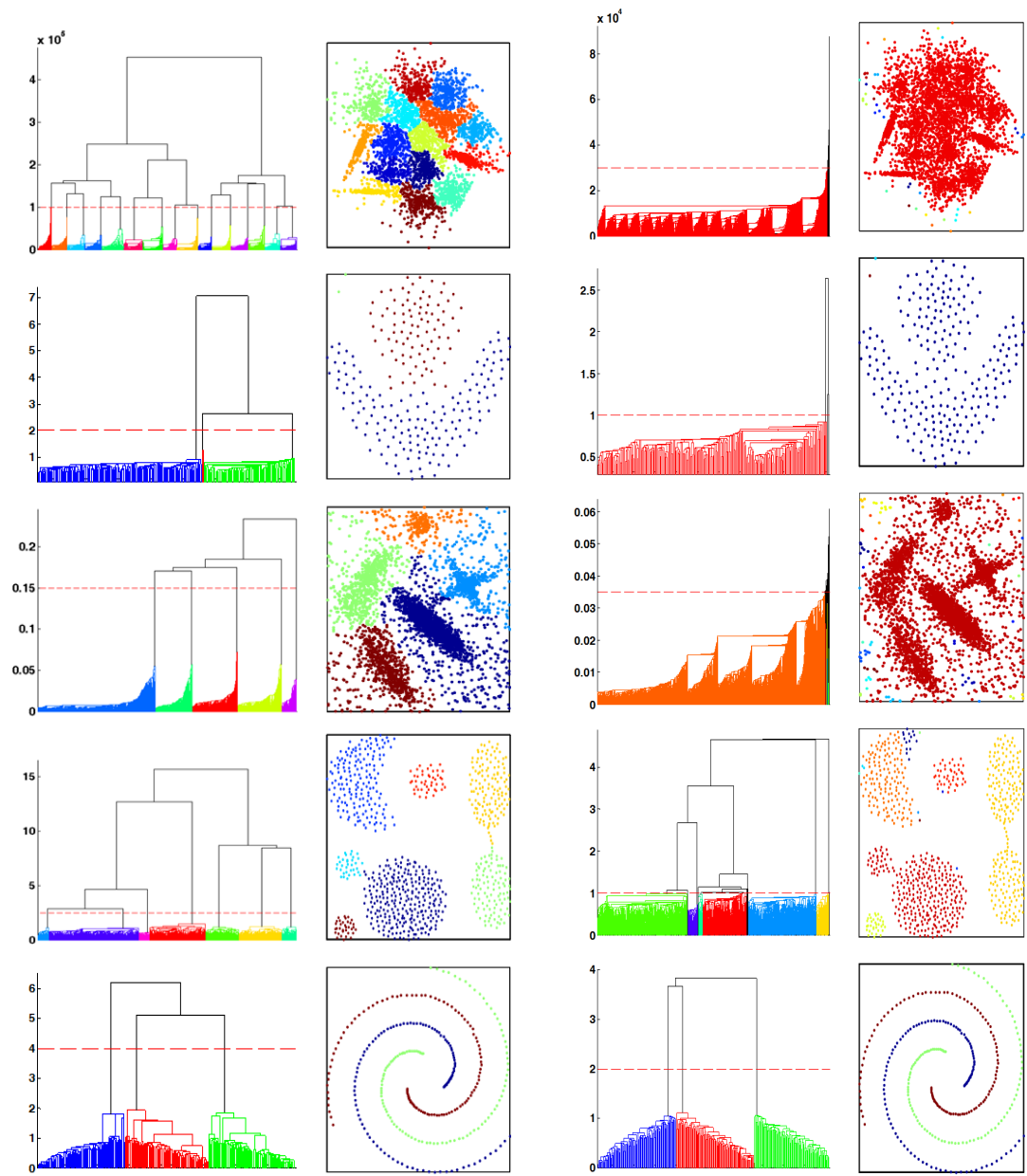

**Fig. 3. Comparison between IT-Dendrogram (the left two columns) and SLHC (the right two columns).** The 1st (leftmost) and 3rd columns show the Dendrograms produced by these two methods. The heights of the red lines denote the thresholds. The 2nd and 4th columns show the corresponding clustering results. Except the last dataset (bottom) with same good performance by the two methods, IT-Dendrogram achieves much better performance than SLHC.

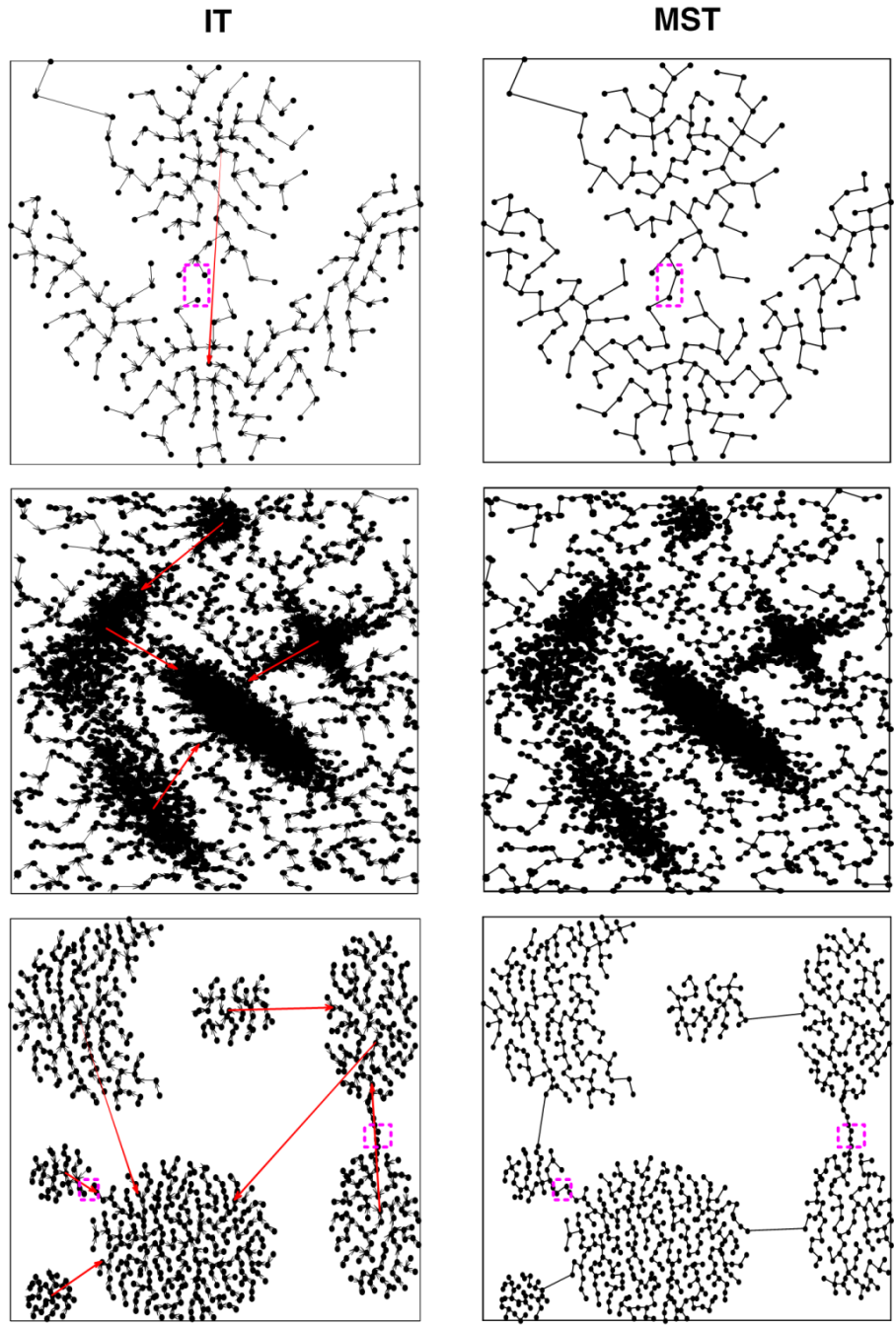

**Fig. 4. Comparison between the IT (left column) and MST (right column) structures for three datasets.** We can see that the undesired connections in the IT structures are much easier to be distinguished and determined, since IT structures usually have long-range connections (in red) between clusters, in contrast to the MST structures with short-range connections especially when two clusters are close to each other or contaminated by noise. Taking the points in red rectangulars for instance, they are connected in the MST, whereas unconnected in the IT structures.

We also tested IT-Dendrogram on high-dimensional datsets (the last two datasets in Table 1) with different attributes. Each data instance in the D32 dataset is a vector with 32 real numbers. Each data intance in the Mushroom dataset[7] is a vector with 22 characters, representing 22 features of a mushroom recorded. The Dendrograms for these two datasets are shown in Fig. 5, in which we can see that some connections are quite inconsistent from others and there are rather large intervals for us to choose the thresholds. The heights of the black dashed lines in Fig. 5 denote the thresholds chosen by us in the experiments. Since these two datasets are all high-dimensional, we cannot directly see the clustering results. However, these two datasets have the annotations that can be used to judge the clustering performance. For instance, each data instance in Mushroom datasets is annotated as poisonous (p) or edible (e). We can thus match the clustering assignments with the annotations and see how consistancy the clustering assignments and those annotations are. The results are excellent. Sixteen clusters are obtained for 32D datasets ($\sigma = 1$) with no error clustering assignment, and 23 clusters for Mushroom dataset ($\sigma = 4$) with no error clustering assignment, either.

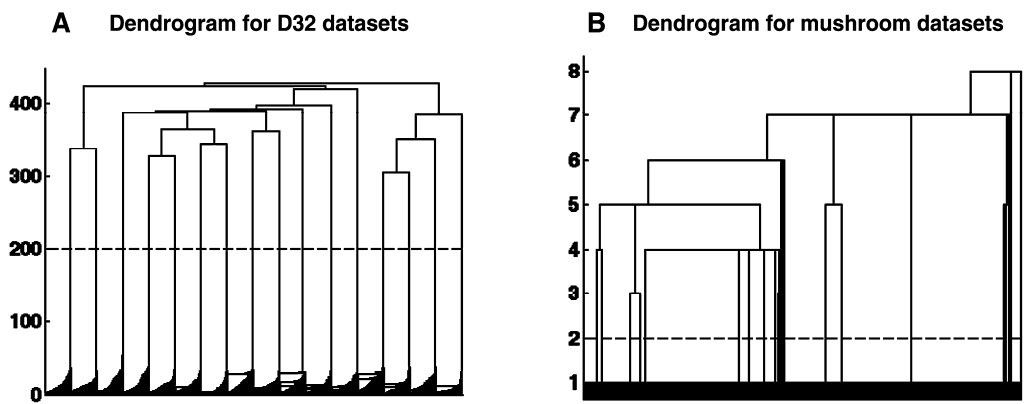

**Fig. 5. Dendrograms of the D32 (left) and Mushroom (right) datasets obtained by IT-Dendrogram.**

**5 Conclusions and discussions**

In this paper, we make an effective combination between the IT structure and the single link hierarchical (SLHC) clustering. As a result, an effective method, called IT-Dendrogram, is proposed to represent the IT structures by the Dendrograms in the low-dimensional space. Our experimental results prove that IT-Dendrogram is another effective visualization method that can help users determine the undesired edges in the IT structures in a visualization environment. Like IT-map, IT-Dendrogram does not require the dimensionality reduction operation for the test datasets.

As shown in Fig. 6A, IT-Dendrogram enriches our IT clustering family. One can also view that IT-Dendrogram enriches the hierarchical clustering methods, too. Centered on the IT structure, there are two main steps in this IT-based clustering family: (i) construct the IT structure and (ii) remove the undesired edges in the IT structure. Strictly speaking, IT-Dendrogram enriches the latter one.

---

[7] Download from http://archive.ics.uci.edu/ml/

Till now, we have tried three different methods to construct the IT structure: NND (*1*), G-NND (2) and N-NND (19, 20). NND refers the initial nearest neighbor descending method. G-NND generalizes the initial NND method to a wider range of types of input, especially the graph-based (the neighborhood graph for instance) or sparse input, and in turn, NND can be viewed as a special case of G-NND. N-NND contains some trials that intend to make the NND method become nonparametric, whereas not satisfactory.

Also, we have proposed a set of methods available for users to remove the undesired edges in the IT structure. These methods are mainly of two categories (Fig. 6C): automatic or interactive. Till now, there are two automatic methods as G-AP (or called IT-AP) (4), IT-SS (*1*), and five interactive methods as IT-Kcut[8] (*1*), IT-Int (*1*), IT-DC (*1*), IT-map (5) and the proposed method IT-Dendrogram. See also Section 1.2 for a description and Fig. 6E for several examples. Note that, for all these automatic and interactive methods, only IT-Int needs the preprocessing (dimensionality reduction) when dealing with the datasets that do not lie in the 2D Euclidean space.

And some discussions for the interactive (or visualization) methods are as follows:

(i) Unlike IT-Kcut and IT-DG which do some feature mapping, IT-Dendrogram and IT-map are the ones that do the structure mapping, which can provide us more information of the IT data structure and thus help users to make more reliable and meaningful cluster analysis on the datasets, especially in some scientific fields such as bioinformatics.

(ii) Compared with IT-map, IT-Dendrogram can always avoid generating the crowding problem, which is useful for users to refer to the nodes in certain problems. However, IT-map is the only method that preserves in the embeddings all the information (i.e., the vectors *I*, *W*, *P*) of the IT data structure, in comparison to the cases for other interactive methods (e.g., IT-Kcut: *W*; IT-DC: *W* and *P*; IT-Dendrogram: *I* and *W*). However, IT-map is also the most complex one (IT-Kcut is the least complex one), too.

(iii) Like IT-map and G-AP, one interesting thing for IT-Dendrogram is that it is generally superior to the method it combines with (i.e., SLHC), which is denoted as IT-Dendrogram ( IT + SLHC ) ≥ SLHC in Fig. 6D.

---

[8] IT-Kcut chooses to cut the edges with the *K* largest dissimilarities. *K* can be judged from the plot regarding the edge dissimilarity.

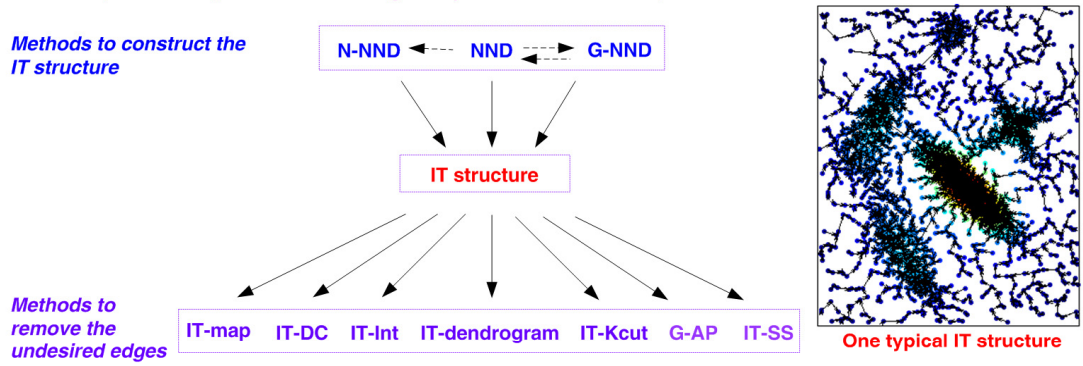

**A: Members (or methods) in the IT clustering family and their relationships**

*Methods to construct the IT structure*: N-NND ← NND → G-NND

→ IT structure

*Methods to remove the undesired edges*: IT-map, IT-DC, IT-Int, IT-dendrogram, IT-Kcut, G-AP, IT-SS

**One typical IT structure**

**B: Sources for the methods:**
- **IT-map:** Qiu, Li arXiv (1501.06450)
- **IT-dendrogram:** Qiu, Li arXiv (?)
- **NND, IT-Int, IT-Kcut, IT-SS, IT-DC:** Qiu, etc. arXiv (1412.5902)
- **G-AP:** Qiu, Li arXiv (1501.04318)
- **G-NND:** Qiu, Li arXiv (1506.06068)
- **N-NND:** Qiu, Li arXiv (1502.04837, 1502.04502)

**C: Categories for the edge-removing methods:**
- **Interactive:** IT-map, IT-dendrogram, IT-DC, IT-Int, IT-Kcut
- **Automatic:** G-AP, IT-SS

**D: Interesting features in clustering performance**
- IT-map ( IT + Isomap ) $\geq$ Isomap
- G-AP ( IT + AP ) $\geq$ AP
- IT-DC ( IT + DC ) $\approx$ DC
- IT-dendrogram ( IT + SLHC ) $\geq$ SLHC

**Notes:**
- Isomap: Tenenbaum, etc. (Science-2000)
- AP: Frey and Dueck (Science-2007)
- DC: Rodriguez and Laio (Science-2014)

**E: Several examples**

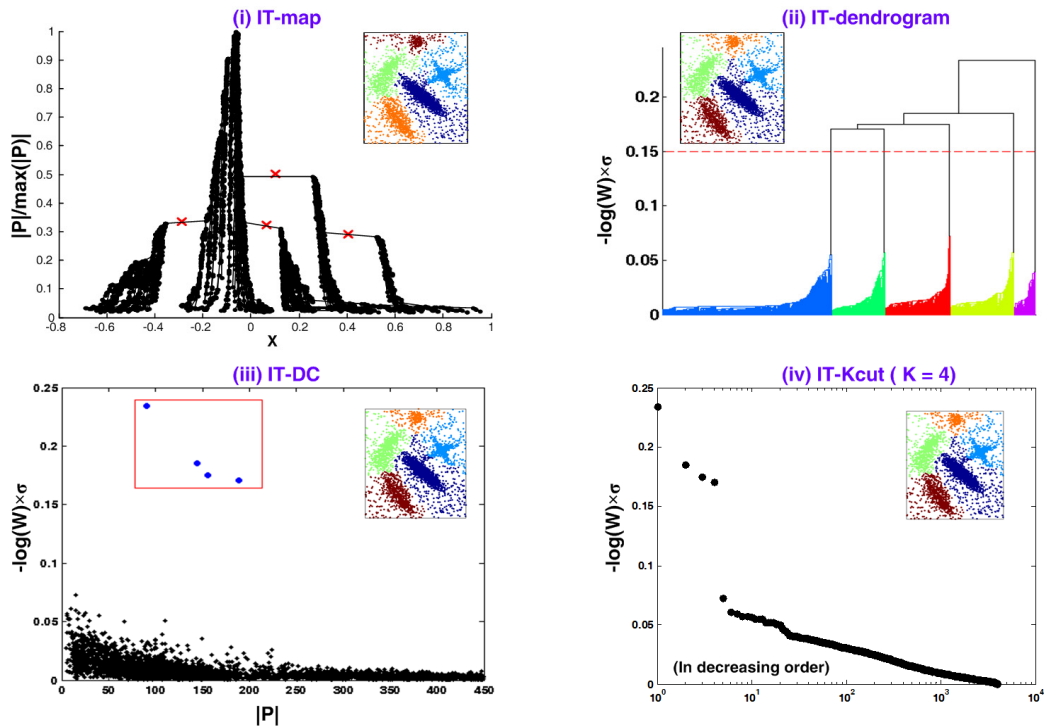

(i) IT-map  (ii) IT-dendrogram  (iii) IT-DC  (iv) IT-Kcut ( K = 4 )

**Fig. 6. An overview of the IT clustering family**

# References

1. Qiu T, Yang K, Li C, & Li Y (2014) A Physically Inspired Clustering Algorithm: to Evolve Like Particles. *arXiv preprint arXiv:1412.5902*.
2. Qiu T & Li Y (2015) A general framework for the IT-based clustering methods. *arXiv preprint arXiv:1506.06068*.
3. Zahn CT (1971) Graph-theoretical methods for detecting and describing gestalt clusters. *IEEE Trans. Comput.* 100(1):68-86.
4. Qiu T & Li Y (2015) A Generalized Affinity Propagation Clustering Algorithm for Nonspherical Cluster Discovery. *arXiv preprint arXiv:1501.04318*.
5. Qiu T & Li Y (2015) IT-map: an Effective Nonlinear Dimensionality Reduction Method for Interactive Clustering. *arXiv preprint arXiv:1501.06450*.
6. Hotelling H (1933) Analysis of a complex of statistical variables into principal components. *Journal of educational psychology* 24(6):417.
7. Torgerson WS (1952) Multidimensional scaling: I. Theory and method. *Psychometrika* 17(4):401-419.
8. Tenenbaum JB, De Silva V, & Langford JC (2000) A global geometric framework for nonlinear dimensionality reduction. *Science* 290(5500):2319-2323.
9. Roweis ST & Saul LK (2000) Nonlinear dimensionality reduction by locally linear embedding. *Science* 290(5500):2323-2326.
10. Van der Maaten L & Hinton G (2008) Visualizing data using t-SNE. *Journal of Machine Learning Research* 9(2579-2605):85.
11. Rodriguez A & Laio A (2014) Clustering by fast search and find of density peaks. *Science* 344(6191):1492-1496.
12. Theodoridis S & Koutroumbas K (2009) *Pattern Recognition, Fourth Edition* (Academic Press Elsevier).
13. Eisen MB, Spellman PT, Brown PO, & Botstein D (1998) Cluster analysis and display of genome-wide expression patterns. *Proc. Natl. Acad. Sci. U.S.A.* 95(25):14863-14868.
14. Fränti P & Virmajoki O (2006) Iterative shrinking method for clustering problems. *Pattern Recognit.* 39(5):761-775.
15. Gionis A, Mannila H, & Tsaparas P (2007) Clustering aggregation. *ACM Trans. Knowl. Discovery Data* 1(1):4.
16. Fu L & Medico E (2007) FLAME, a novel fuzzy clustering method for the analysis of DNA microarray data. *BMC Bioinf.* 8(1):3.
17. Chang H & Yeung D-Y (2008) Robust path-based spectral clustering. *Pattern Recognit.* 41(1):191-203.
18. Franti P, Virmajoki O, & Hautamaki V (2006) Fast agglomerative clustering using a k-nearest neighbor graph. *Pattern Analysis and Machine Intelligence, IEEE Transactions on* 28(11):1875-1881.
19. Qiu T & Li Y (2015) Clustering by Descending to the Nearest Neighbor in the Delaunay Graph Space. *arXiv preprint arXiv:1502.04502*.
20. Qiu T & Li Y (2015) Nonparametric Nearest Neighbor Descent Clustering based on Delaunay Triangulation. *arXiv preprint arXiv:1502.04837*.